\pdfoutput=1

\documentclass[11pt]{article}

\usepackage[final]{acl}

\usepackage{times}
\usepackage{latexsym}

\usepackage[T1]{fontenc}

\usepackage[utf8]{inputenc}

\usepackage{microtype}

\usepackage{inconsolata}

\usepackage{graphicx}
\usepackage[inline]{enumitem}

\title{Systematic Evaluation of Machine-Generated Reasoning and PHQ-9 Labeling for Depression Detection Using Large Language Models}

\author{Zongru Shao \\ Silicon Austria Labs \\ \small{\texttt{zongru.shao@silicon-austria.com}} \And
         Xin Wang \\ Jiangnan University \\ \small{\texttt{wxlboro@jiangnan.edu.cn}} \And
         Zhanyang Liu \\ Jiangnan University \\ \small{\texttt{6243112035@stu.jiangnan.edu.cn}}
         \AND
         Chenhan Wang \\ Jiangnan University \\ \small{\texttt{1033230224@stu.jiangnan.edu.cn}} \And
         K.P. Subbalakshmi \\ Stevens Institute of Technology \\ \small{\texttt{ksubbala@stevens.edu}}  }

\begin{document}
\maketitle
\begin{abstract}
Recent research leverages large language models (LLMs) for early mental health detection, such as depression, often optimized with machine-generated data. 
However, their detection may be subject to unknown statistical biases which imposed detection inaccuracy. Meanwhile, quality control has not been applied to these generated corpora besides sampled human verifications, which is significantly limited compared to the scale of the data in use. 
Our goal in this work is to systematically evaluate LLM reasoning and reveal potential statistical biases.
To this end, we first provide a systematic evaluation of the reasoning over machine-generated detection and interpretation, thus revealing potential statistic detection biases in the context. Then we use the models' reasoning abilities to explore mitigation strategies for enhanced performance. 
Specifically, we do the following: 
\begin{enumerate*}[label=(\alph*)] 
\item Design an LLM instruction strategy that allows for systematic analysis of the classification by breaking down the detection task into several subtasks. 
\item Design contrastive few-shot and chain-of-thought prompts by selecting typical positive and negative examples of detection reasoning.
\item Perform human annotation for the subtasks identified in the first step and evaluate the few-shot performance. 
\item Identify human-preferred detection with desired logical reasoning from the few-shot generation and use them to explore different optimization strategies. 
\end{enumerate*} 
We conducted extensive comparisons of the experimental results on the DepTweet dataset across the following subtasks: 
\begin{enumerate*}[label=(\arabic*)] 
\item identifying whether the speaker is describing their own depression, 
\item accurately detecting the presence of PHQ-9 symptoms, and
\item finally, detecting depression. 
\end{enumerate*} 
Human verification of statistical outliers shows that LLMs demonstrate greater accuracy in analyzing and detecting explicit language of depression as opposed to implicit expressions of depression.
We also note that, the LLMs are biased towards making a ``depression'' decision when there are explicit depression-related keywords. By contrast, they are biased towards ``non-depression" decisions, when there is no depression-related keyword in the text.
Two optimization methods are used for performance enhancement and reduction of the statistic bias: (1) supervised fine-tuning (SFT) and (2) direct preference optimization (DPO). Notably, the DPO approach achieves significant performance improvement.
\end{abstract}

\section{Introduction}
\label{sec:intro}

It is estimated that approximately 5\% of adults worldwide suffer from depression\footnote{WHO (World Health Organization), \url{https://www.who.int/news-room/fact-sheets/detail/depression}}, which not only negatively impacts their emotions, behaviors, and cognitive processes but also potentially leads to self-harm or even suicide. Thus, early detection and diagnosis are crucial for effective treatment to minimize mortality and morbidity~\cite{goldman1999awareness}. 
Social media platforms provide valuable insights on mental health, complementing traditional clinical diagnostics, especially with advances in natural language processing (NLP) \cite{de2013predicting}.
Traditional NLP approaches detect depression through linguistic features like word frequencies and emotional keywords \cite{lyu2023detecting} while dismissing complex and indirect expressions. The emergence of transformer-based \cite{vaswani2017attention} language models has enabled nuanced analysis of expression interactions within the context~\cite{zogan2023hierarchical,wang2021modular} to improve the detection performance, but they are supervised by the training data and fail to generalize for unseen scenarios \cite{harrigian2020models}. 
Recent advancements of LLMs have significantly impacted the automated screening of mental health issues~\cite{xu2023leveraging}.
Several approaches fine-tuned open LLMs for this purpose, such as \textit{MentaLLaMA}, \textit{MentalBART}, \textit{MentalT5} \cite{yang2023mentalllama}, while their interpretability of reasoning was facilitated by a training dataset generated by proprietary LLMs.
Meanwhile, SEGA \cite{chen2024depression} transformed clinical interviews into expertise-driven graphs and leveraged LLMs for data augmentation, significantly boosting performance on clinical datasets. 
Diagnostic reasoning achieved by LLMs \cite{savage2024diagnostic} has garnered significant interest in related studies. However, comprehensive evaluations of the generated rationales are still conducted by manual assessment of physicians and medical experts \cite{savage2024diagnostic}, limiting the volume of throughput in their analysis.
We want to automate the evaluation of reasoning so that the LLM-generated decisions and associated rationales can be evaluated in large quantities. Therefore, we propose a breakdown of the depression detection task into several subtasks and explore a few methods to use the LLM-generated rationales to improve overall system performance.  
The objectives of this work are as follows:
\begin{enumerate*}[label=(\alph*)] 
\item To systematically evaluate MentalLLM-generated detection in terms of their adherence to instructions/prompts and their reasoning. 
\item To identify potential statistical biases in the decision-making compared to human judgments.
\item To explore the utilization of generated reasoning from open-source LLMs for further model optimization. 
\end{enumerate*}
Consequently, we design an LLM-based depression-detection system to identify a range of subtasks and evaluate their responses with subtask-labeling annotated by human experts. The analyses provide insights into effective instruction tuning of LLMs with machine-generated data, as well as an understanding of the potential weaknesses in LLM-based depression detection.

\section{Defining Subtasks for Depression Detection}

Language models have attracted research interests for text-based depression detection, such as \textit{MentaLLaMA} and other related works.
However, the quality of the generated reasoning in the context is evaluated by human verification, which is costly and unscalable. Current AI systems still struggle to perform rigorous automated logical evaluation from the flow of the text. To address these challenges, we decompose the overall task of depression detection from tweet text into several subtasks that act as critical ``checkpoints'' within the logical reasoning process of the LLMs. This enables us to conduct a systematic analysis of the process, the outcome, and potential weaknesses.

Implicit expressions and circumlocutory language pose significant challenges for traditional NLP approaches \cite{despot2023somewhere}. This is often caused by the fact that the presence or absence of specific keywords can disproportionately impact the analysis. In particular, we identify two key scenarios where breaking down the depression prediction task into subtasks is especially beneficial:
\begin{enumerate*}[label=(\alph*)] 
\item When depression-related keywords are present but an analysis of the linguistic context demonstrates that the text does not describe the speaker's own depressive state (for instance, when the text pertains to general knowledge or refers to another individual).
\item When depression keywords are absent, yet the speaker implicitly conveys extremely negative emotions, suggesting a high likelihood of depression as estimated by human experts.
\end{enumerate*}

The presence of depressive symptoms is another key factor for depression detection. The Patient Health Questionnaire (PHQ-9) \cite{kroenke2001phq} is a widely accepted self-report tool used to screen for both the presence and severity of depression. Given its simplicity and significant domain contribution, we leverage the PHQ-9 framework to break down our task as follows:
\begin{enumerate*}[label=(\arabic*)] 
\item \textbf{Self-reference Analysis}: Determine whether the \textbf{S}peaker is describing their own mental state (S = \textit{Yes} or \textit{No}).
\item \textbf{Symptom Detection}: Evaluate the presence of each of the PHQ-9 symptoms (S1–S9, as described in \cite{kabir2023deptweet}).
\item \textbf{Overall Diagnosis}: Make a final decision regarding \textbf{D}epression (D) by synthesizing the analyses from the previous steps.
\end{enumerate*}
In total, depression detection from tweet text is decomposed into eleven subtasks: one for self-reference identification, nine for the PHQ-9 symptoms, and the last for the final depression diagnosis.

\section{LLM Learning \& Tuning Framework}
\label{sec:method}

\begin{figure*}[!h]
\centering
\includegraphics[width=0.99\textwidth]{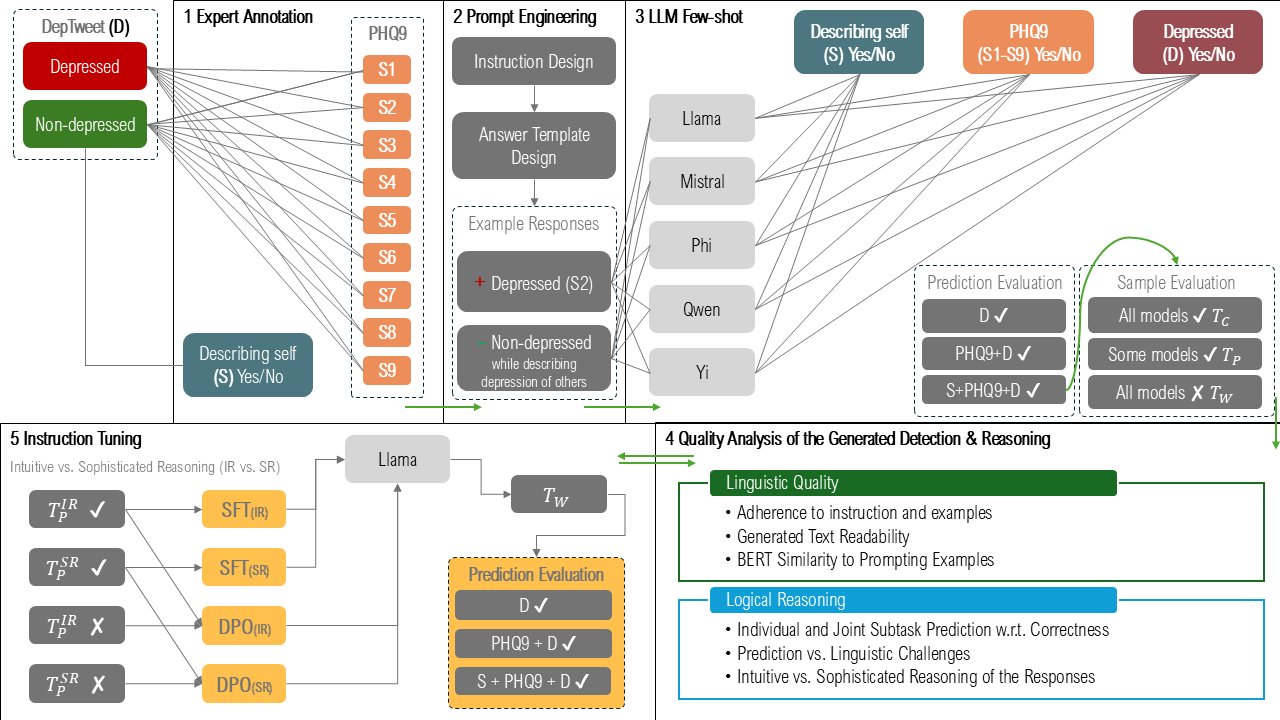}
\captionsetup{font=footnotesize}
\caption{An overview of the LLM-based detection and analysis framework.}
\label{fig:pipeline}
\end{figure*}

We design a comprehensive experimental framework consisting of five main components, as illustrated in Figure \ref{fig:pipeline}:
\begin{enumerate*}[label=\arabic*).] 
\item \textbf{Expert Annotation}: Establishes human-derived references to guide the creation of appropriate subtasks.
\item \textbf{Prompt Engineering}: Develops a detailed instruction set and provides two prompting examples (one positive and one negative) to differentiate between the target cases.
\item \textbf{LLM Few-shot}: Leverages LLMs to analyze and generate predictions for all subtasks based on given text inputs.
\item \textbf{Quality Analysis of the Generated Detection \& Reasoning}: Evaluates the generated analysis from both linguistics and logical reasoning perspectives and ensure the quality of the content.
\item \textbf{Instruction Tuning}: Refines the LLMs using different optimization strategies with desired analysis from the few-shot generation.
\end{enumerate*}
Each of these components is described in detail in the following subsections.

\subsection{Expert Annotation}
\label{subsec:annotation}

We employ the Deptweet dataset as our benchmark due to its strong alignment with established clinical frameworks (DSM-5, PHQ-9). This dataset comprises 40,191 tweets that have been expertly annotated with depression severity and the annotator confidence of their labeling. To reduce annotation ambiguity where the tweet lacks relevant context and it is insufficient to derive a \textit{depressed} or \textit{non-depressed} label, we select 1,566 tweets where the experts were confident enough with their \textit{depressed} annotations (confidence > 0.95). We randomly select another 1,566 samples where the experts were confident enough with their \textit{non-depressed} annotations, resulting in a balanced dataset ($N = 3,132$). We disregard the depression severity to further reduce potential annotation disagreement, leaving only \textit{depressed} or \textit{non-depressed}. Then, we perform our annotation with self-reference (\textbf{S}) and the PHQ-9 symptoms (\textbf{S1}–\textbf{S9}) in a binary format (\textit{Yes} or \textit{No}), following the protocol outlined in \cite{kabir2023deptweet}. The annotation process was performed by three psychology professionals and three graduate students trained to identify PHQ-9 related keywords, underlying causes, and self-reference of the speakers.

\subsection{Prompt Engineering}
\label{subsec:prompt}

\subsubsection{Instruction}

Prompt engineering is critical for effective LLM-based solutions \cite{antunes2023prompting}. Our instruction set integrates Chain-of-Thought (CoT) reasoning \cite{chu2023survey} and comprises four essential elements:
\begin{enumerate*}[label=(\alph*)] 
\item An explicit definition of the agent’s role as an experienced clinical professional.
\item A high-level description of the overall task that emphasizes key domain knowledge.
\item A detailed breakdown of the subtasks, including their descriptions, annotation labels, and underlying reasoning.
\item Quality assurance mechanisms \cite{widyasari2024beyond} to facilitate self-checks and prevent errors.
\end{enumerate*}
We iteratively refine the instruction with LLM feedback to ensure its precised description.

\subsubsection{Prompting Examples}

To validate our instructions, we develop two prompting examples that follow the defined steps. These examples are refined with LLM-generated feedback to ensure their quality and clarity. One example represents a positive case by exhibiting the \textit{feeling down} symptom (S2), while the other represents a negative case by including the term “depression” in a context that describes someone else. This dual approach ensures that our prompt examples capture both typical and challenging scenarios.

\subsection{LLM Few-shot}
\label{subsec:fewshot}

For the few-shot learning experiments, we use open-source 7B-scale LLMs known for their GPU efficiency and reliable reasoning capabilities, as evidenced by benchmarks such as MMLU-Pro \cite{wang2024mmlu}. Given that our depression detection process is chat-based, we prioritize instruction-tuned models that support role-play (e.g., through a customizable \textit{system} role). Table \ref{tab:llm} lists the five flagship models selected for our experiments along with their latest versions.

\begin{table}[!htbp]
\small
  \centering
  \begin{tabular}{l|r}
    \hline
    LLM & Version \\
    \hline
    Llama & Llama-3.1-8B-Instruct \\
    Mistral & Mistral-Nemo-Instruct-2407  \\
    Phi & Phi-3.5-mini-instruct  \\ 
    Qwen & Qwen2.5-7B-Instruct  \\ 
    Yi & Yi-1.5-9B-Chat  \\ 
    \hline
  \end{tabular}
  \captionsetup{font=footnotesize}
  \caption{Selected LLMs in our experiment.}
  \label{tab:llm}
\end{table}

\subsection{Quality Analysis of Generated Diagnoses}
\label{subsec:quality}

Both linguistic quality and logical reasoning are considered for quality analysis of the generated analysis of the detection, focusing on their closeness with the prompting examples.

\subsubsection{Linguistic Quality}
We evaluate the generated text using several metrics that capture factual content and adherence to expected formats:
\begin{enumerate*}[label=\alph*)] 
\item \textit{Adherence to the format (A)}: Whether the output follows the specified response format (as exemplified in the prompting examples) so that the logical flow is close to the human analysis in the prompting examples while the annotation labels can be automatically extracted.
\item \textit{Automated Readability Index (ARI)} \cite{smith1967automated}: Whether the text is as readable as the prompting examples.
\item \textit{BERT Textual Similarity}: The semantic closeness between the generated output and the prompting examples, as determined by BERT embeddings which capture both phrase-level and hierarchical linguistic features \cite{jawahar2019does}.
\end{enumerate*}
The format adherence metric is defined as:
\begin{equation}
A = \frac{N_{a}}{N}
\end{equation}
where $N$ is the total number of input samples, and $N_{a}$ is the number of responses that fully adhere to the format (i.e., all subtask labels). These metrics are analyzed statistically, and outlier responses are further reviewed manually to ensure quality.

\subsubsection{Logical Reasoning}
The overall reasoning quality of the generated responses is assessed from three perspectives:
\begin{enumerate*}[label=\arabic*)] 
\item \textbf{Accuracy}: Whether each individual subtask is correctly predicted and whether the collective analysis aligns with human judgment.
\item \textbf{Weaknesses Assessment}: Identification of potential weaknesses, particularly with respect to challenges in:
    \begin{enumerate*}[label=(\alph*)]
    \item self-references of the speakers' depressive states and other experiences (vs. describing depressive states of others) and
    \item the influence of explicit mention of depression-related keywords.
    \end{enumerate*}
\item \textbf{Comparative Analysis}: Compare different reasoning schemes w.r.t. the sequence of the key subtask ``checkpoints'', which is based on human observations of the generated diagnoses.
\end{enumerate*}

To evaluate the performance of the overall joint decision-making for depression detection, we define the \textit{correct ratio} $C$ as:
\begin{equation}
C = \frac{N_{c}}{N}
\end{equation}
where $N_{c}$ represents the number of samples with fully correct responses, and $N$ is the total number of input samples.

\subsection{Instruction Tuning}
\label{subsec:tuning}

To further improve logical reasoning of the LLMs, we perform instruction tuning using high-quality diagnoses generated by few-shot. We implement two optimization strategies:
\begin{enumerate*}[label=(\alph*)] 
\item \textbf{Supervised Fine-Tuning (SFT)}: Following state-of-the-art approaches such as \textit{MentaLLaMA}.
\item \textbf{Preference-Based Optimization}: Specifically, Direct Preference Optimization (DPO), which has demonstrated improved performance over SFT on standard benchmarks \cite{rafailov2024direct, ivison2023camels}.
\end{enumerate*}
Our experiments compare these strategies under limited computational resources by tuning an LLM on the few-shot detection and reasoning with their generated responses.

Given $K$ LLMs in our few-shot setting (denoted by $LLM_1$, $LLM_2$, $\dots$, $LLM_K$), each input sample receives $K$ detection responses. We label a response as \textit{correct} (\textbf{C}) if it accurately predicts all eleven subtasks. Accordingly, we classify each sample as:
\begin{itemize*}
  \item \textbf{Easy}: All $K$ responses are correct.
  \item \textbf{Hard}: All $K$ responses are wrong (\textbf{W}).
  \item \textbf{Partially Correct}: Some responses are correct while others are not.
\end{itemize*}
Thus, the overall dataset of $N$ samples is partitioned into three collections:
\begin{enumerate*}[label=(\alph*)] 
\item the all-\textit{correct} collection $T_{C}$,
\item the \textit{partially correct} collection $T_{P}$, and
\item the all-\textit{wrong} collection $T_{W}$.
\end{enumerate*}
Correspondingly, the generated responses ($R$) are divided into four groups: 
\begin{enumerate*}[label=(\arabic*)] 
  \item Correct responses from $T_C$: $\{ R^{C}_{i,j} \mid i \in T_C, \, j \in \{1,2,\dots,K\} \}$.
  \item Correct responses from $T_P$: $\{ R^{C}_{i,j} \mid i \in T_P \}$.
  \item Wrong responses from $T_P$: $\{ R^{W}_{i,j} \mid i \in T_P \}$.
  \item Wrong responses from $T_W$: $\{ R^{W}_{i,j} \mid i \in T_W, \, j \in \{1,2,\dots,K\} \}$.
\end{enumerate*}
For SFT, we use the correct responses from the partially correct set ($\{ R^{C}_{i,j} \mid i \in T_P \}$). For DPO, we form training pairs consisting of correct and wrong responses from $T_P$, and we evaluate performance improvements on the more challenging hard set $T_{W}$.

\section{Experimental Results}

The analysis of the experimental results is conducted from several perspectives: 
\begin{enumerate*}[label=(\arabic*)]
\item Statistical distribution of the expert annotations,
\item Few-shot performance compared to state-of-the-art Mental-LLMs with human verifications from various aspects considering potential detection weaknesses,
\item A thorough comparison of instruction tuning schemes, and
\item Performance differences w.r.t. different sample characteristics -- self-reference of the speakers and the presence of depression-related keywords.
\end{enumerate*}

\subsection{Statistical distribution of the expert annotations}

We first analyze the distribution of human annotations in DepTweet, noting that all annotations are binary with \textit{Yes}/\textit{No} indications. The distribution of the annotations is shown in Table \ref{table:annotation}.
To consider \textit{explicit} vs. \textit{implicit expressions} of depression, we extract the depression-related keywords (e.g., ``depress'', ``depressant'', ``depressed'', ``depressing'', ``depression'', ``depressive'', etc.) and separate the samples into two groups: \textit{Mentioned Depression (MD)} and \textit{No Mention of Depression (NMD)}.
Table \ref{table:annotation} reveals that eight out of the nine PHQ-9 symptoms are imbalanced with significantly less occurrences, accompanied by the exception of \textbf{S2} (\textit{feeling down}). 
The majority of the depressed samples contain depression-related keywords, while the non-depressed samples do not. However, the presence of such keywords cannot be overlooked in the non-depressed group, which constitutes about 22\% of the samples. Additionally, 27\% of the depressed samples do not contain these keywords, indicating implicit expressions of depression. 
Given the imbalanced nature of these subtask annotations in the dataset, we use the micro F1 score \cite{chicco2020advantages} to evaluate single-label detection and the \textit{correct ratio} $C$ for joint decision-making of multiple subtasks.

\begin{table}[!htbp] \scriptsize
\centering
\begin{tabular}{l||r|r||r|r|r|r}
\hline
\textbf{D} (DepTweet) & \multicolumn{2}{c||}{\textit{Depressed} (\%)} & \multicolumn{4}{c}{\textit{Non-depressed} (\%)} \\ \hline
\textbf{S} (ours) & \multicolumn{2}{c||}{Yes} & \multicolumn{2}{c|}{No} & \multicolumn{2}{c}{Yes} \\ \hline
Annotation & No & Yes & No & Yes & No & Yes \\ \hline
\textbf{S1} (ours)         & 96.4 & 3.6  & 26.8 & 0    & 73.2 &  0 \\
\textbf{S2} (ours)         & 7.9  & 92.1 & 25.8 & 1.0  & 63.8 & 9.4 \\
\textbf{S3} (ours)         & 94.8 & 5.2  & 26.8 & 0.1  & 72.0 & 1.1 \\
\textbf{S4} (ours)         & 73.1 & 26.9 & 25.6 & 1.2  & 55.6 & 17.6 \\
\textbf{S5} (ours)         & 97.3 & 2.7  & 26.8 & 0    & 73.0 & 0.2 \\
\textbf{S6} (ours)         & 87.8 & 12.2 & 26.8 & 0    & 72.7 & 0.5 \\
\textbf{S7} (ours)         & 98.7 & 1.3  & 26.8 & 0    & 73.2 & 0 \\
\textbf{S8} (ours)         & 95.7 & 4.3  & 26.8 & 0.1  & 72.7 & 0.5 \\
\textbf{S9} (ours)         & 85.1 & 14.9 & 26.8 & 0    & 73.1 & 0.1 \\ \hline
\textit{NMD}/\textit{MD}   & 26.8 & 73.2 & 19.3 & 7.5  & 58.8 & 14.4 \\ 
\hline
\end{tabular}
\captionsetup{font=footnotesize}
\caption{Distribution of 11 subtask annotations, \textbf{\{S, S1, S2, ..., S9, and D\}}: percentage in the \textit{Depressed} or \textit{Non-depressed} groups; \textit{MD} refers \textit{Mentioned Depression}; \textit{NMD} denotes \textit{No Mention of Depression}; ``Annotation'' indicates the human annotation of the PHQ-9 symptoms in binary \textit{Yes} or \textit{No} form and the classification of the \textit{MD} (\textit{Yes}) or \textit{NMD} (\textit{No}) group.).}
\label{table:annotation}
\end{table}

\subsection{State-of-the-art Mental-LLMs}

\begin{figure}[!h]
\centering
\includegraphics[width=0.99\columnwidth]{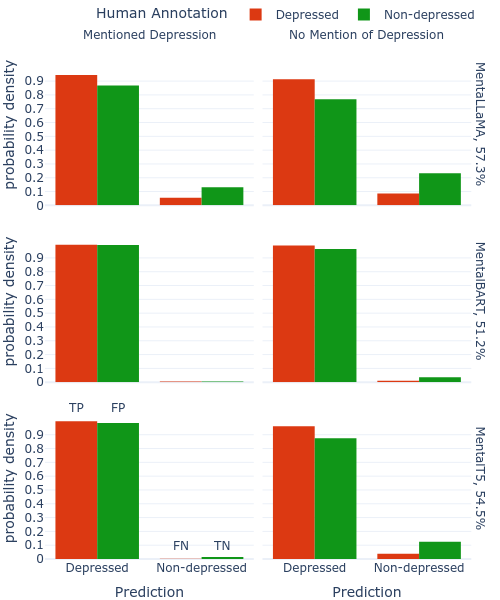}
\captionsetup{font=footnotesize}
\caption{Evaluation of state-of-the-art Mental-LLMs with F1 scores (formatted as ``\textit{LLM}, \textit{F1}\%''). $TP$ denotes true positive, $FP$ false positive, $FN$ false negative, and $TN$ true negative. A higher $FP$ in the left column (\textit{Mentioned Depression (MD)}) compared to the right (\textit{No Mention of Depression (NMD)}) indicates that ``the model has a tendency to consider the samples as depressed when these keywords are present''.}
\label{fig:sota-eval-md-nmd}
\end{figure}

We evaluate finetuned Mental-LLMs from prior works \cite{yang2023mentalllama}, including \textit{MentalLlama}, \textit{MentalBART}, and \textit{MentalT5}. Figure \ref{fig:sota-eval-md-nmd} illustrates their performance w.r.t. the \textit{Depressed}, \textit{Non-depressed}, \textit{Mentioned Depression (MD)}, and \textit{No Mention of Depression (NMD)} groups. Notably, none of these models were fine-tuned on DepTweet. The results indicate that all three models exhibit poor performance, with low accuracy, as the majority of non-depressed tweets are misclassified. 
There could be two reasons for such poor performance: 
\begin{enumerate*}[label=(\arabic*)] 
\item These models were not optimized with DepTweet, which also implies poor generalization.
\item These models were finetuned with data generated by proprietary LLMs, which might introduce unknown deviations and biases.
\end{enumerate*}
Although their generated responses facilitate reasoning, these models do not analyze PhQ-9 symptoms. Thus, we only evaluate their detection of depression.
We note that the false positive (FP) rate is lower in cases when depression related keywords are present, but the false negative (FN) rate is higher when these keywords are absent. This indicates that the models have an apparent tendency to consider the samples as depressed when these keywords are present, while they tend to recognize the samples as non-depressed when these keywords are absent amid the poor performance.

\subsection{Few-shot Evaluation}

In comparison to the state-of-the-art Mental-LLMs, we conduct a systematic evaluation of the generated detection from two perspectives: their linguistic quality and logical reasoning.

\subsubsection{Linguistic Quality}
We use several automated metrics indicating linguistic quality of the generated responses to evaluate their closeness to the prompting examples (as references):
\begin{enumerate*}[label=(\alph*)] 
\item Generation length and \textit{adherence to the format} (\textbf{A} is assessed in Table \ref{table:few-shot}), 
\item \textit{Automated Readability Index (ARI)}, and 
\item Average cosine \textit{BERT similarity} with the input examples.
\end{enumerate*}
The generation length has a mean of 725 tokens with a standard deviation of 135 (denoted as 735$\pm$135, noting that the two prompting examples are 789 and 685 tokens in length, respectively, as references). The readability (ARI) is 22.2$\pm$2.3 (with reference values of 23.0 and 23.9), and the BERT similarity is 0.98$\pm$0.03. 
These metrics show that most of the generated responses align with the prompting examples from these linguistic aspects. Human verification is applied to sampled responses, particularly for the outliers, to confirm their correctness and to observe behaviors of the tested models.  
It yields several observations from human reads regarding the generated detection using few-shot learning:
\begin{enumerate*}[label=(\arabic*)]
\item \textbf{Short generation} refers to short diagnoses that is incomplete (w.r.t. the subtasks). LLMs vary in behavior for short generation: e.g., Llama avoids analyzing extreme emotions, while Mistral generates short responses more often when the text is irrelevant to depression.
\item \textbf{Long responses} can be beneficial when the LLM iteratively analyzes multiple individuals, leading to a longer diagnosis. Additionally, the LLM may revise its predictions, rewriting the reasoning with updated labels.
\item \textbf{Desired response length} does not qualify the generation to be adhering to the prompting examples. Extracting the labels can be challenging even when all subtask annotations are present and intact if the desired format is misplaced.
\item \textbf{Annotation label confusion}. The LLM may occasionally confuse annotation labels, even though the template is implemented as intended.
\end{enumerate*}

\subsubsection{Logical Reasoning}

\begin{table}[!htbp]\small
\centering
\scriptsize
\begin{tabular}{l|l|l|l|l|l}
\hline
LLM                   & Llama         & Mistral       & Phi           & Qwen          & Yi \\ \hline
\textbf{A}            & 94.8          & 65.7          & 77.4          & 96.0          & \textbf{97.9} \\
\textbf{D} (F1)       & 83.6          & \textbf{87.2} & 78.4          & 86.5          & 84.1 \\
\textbf{S} (F1)       & 84.1          & 82.7          & 85.1          & 81.2          & \textbf{85.3} \\
\textbf{S1} (F1)      & \textbf{90.2} & 75.7          & 50.0          & 83.8          & 72.3 \\
\textbf{S2} (F1)      & 80.9          & 80.4          & 77.2          & \textbf{82.0} & 81.9 \\
\textbf{S3} (F1)      & 95.8          & 92.3          & \textbf{96.5} & 96.1          & 95.3 \\
\textbf{S4} (F1)      & \textbf{86.2} & 79.2          & 76.6          & 83.5          & 84.4 \\
\textbf{S5} (F1)      & 98.4          & 98.0          & \textbf{98.8} & 98.7          & 98.1 \\
\textbf{S6} (F1)      & 92.6          & 80.7          & 88.2          & \textbf{94.1} & 93.1 \\
\textbf{S7} (F1)      & 96.3          & 93.9          & 92.1          & 96.8          & \textbf{97.1} \\
\textbf{S8} (F1)      & 95.8          & 91.1          & 96.2          & \textbf{97.1} & 96.2 \\
\textbf{S9} (F1)      & 94.5          & 95.2          & 95.1          & \textbf{97.7} & 97.6 \\ \hline
PHQ9 ($C$)              & \textbf{54.3} & 40.0          & 28.2          & 51.7          & 43.2 \\ 
PHQ9+D ($C$)            & \textbf{51.3} & 37.5          & 26.1          & 49.4          & 40.6 \\
S+PHQ9+D ($C$)          & \textbf{39.5} & 25.2          & 17.8          & 35.3          & 30.6 \\ \hline
\end{tabular}
\captionsetup{font=footnotesize}
\caption{Evaluation of few-shot learning with LLMs (\textbf{A} denotes \textit{adherence to the format}, \textbf{D} depression, and \textbf{S} \textit{speaker reference}). The best-performing LLM is highlighted in bold.}
\label{table:few-shot}
\end{table}

\begin{figure}[!h]
\centering
\includegraphics[width=0.95\columnwidth]{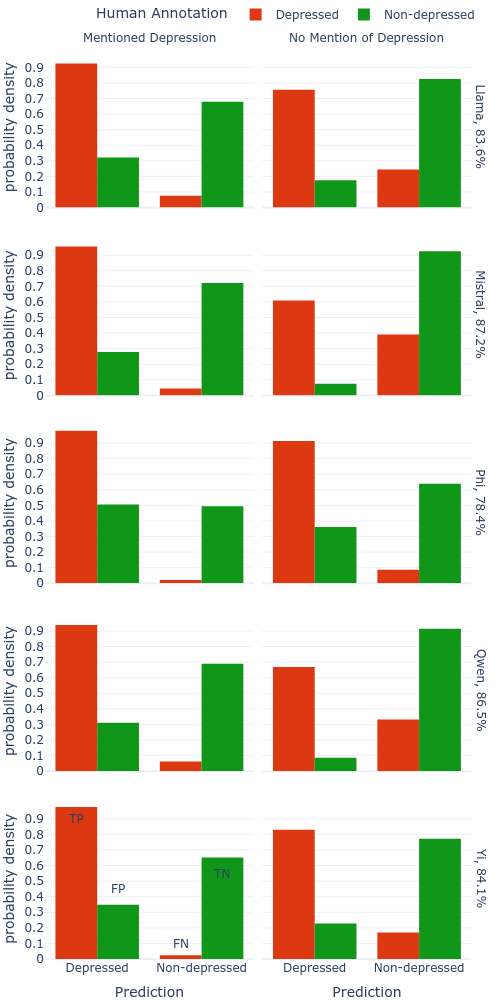}
\captionsetup{font=footnotesize}
\caption{Evaluation of few-shot learning with LLMs w.r.t. different annotation groups (formatted as ``\textit{LLM}, \textit{F1}\%''). $TP$ denotes true positive, $FP$ false positive, $FN$ false negative, and $TN$ true negative.}
\label{fig:fewshot-eval-md-nmd}
\end{figure}

We evaluate logical reasoning based on the alignment of subtask predictions with human annotations. If all subtask predictions match, reasoning is considered ``accurate" and \textit{correct} under a well-designed task breakdown. Subtask predictions and their joint correctness are summarized in Table \ref{table:few-shot}, while the final decision-making is further assessed in Figure \ref{fig:fewshot-eval-md-nmd} w.r.t. different annotation groups.

\textbf{Subtask Predictions.} Table \ref{table:few-shot} demonstrates the following: 
\begin{enumerate*}[label=(\alph*)] 
\item Llama, Qwen, and Yi more frequently adhere to the format compared to Mistral and Phi.
\item All five LLMs outperform state-of-the-art Mental-LLMs in Figure \ref{fig:sota-eval-md-nmd} w.r.t. the detection of depression (\textbf{D}). 
\item Despite the effective detection of subtasks ($>$80\% with Llama), achieving a correct detection encompassing all desired sub-labels remains significantly challenging (e.g., the \textit{correct ratio} $C$ is 39.5\% with Llama compared to an F1 of 83.6\% of depression detection). This indicates that these LLMs are insufficiently capable of providing the comprehensive analysis.
\end{enumerate*}

\textbf{Linguistic challenges} w.r.t. the presence of depression-related keywords. Figure \ref{fig:fewshot-eval-md-nmd} reveals the correlation between the detection of depression and the presence of the depression keywords. It indicates that false positives (FP) are higher when depression keywords are present, and false negatives (FN) are higher when they are absent. Nevertheless, the overall performance is significantly improved compared to state-of-the-art Mental-LLMs. (Note that such a phenomenon is not observed for \textit{speaker self-references}.)

\textbf{Revision of Reasoning.}  It is important to note that some responses provide multiple annotations for the same subtask, potentially with revisions. We specifically examine the flow of logic and correctness in these cases. From the few-shot results, 34 responses with multiple same-task annotations were extracted: 13 generated by Yi, 3 by Qwen, 9 by Llama, and 9 by Mistral. Yi confused the labels and generated the same label twice on 9 occasions. Another common scenario is that the LLM revised its predictions, particularly the \textit{speaker reference}, or analyzed the relevant individuals one by one. Additionally, the LLM may express uncertainty in its reasoning by stating, ``I cannot conclude …'' even when all prediction labels are present.

\subsection{Instruction Tuning}

\subsubsection{Intuitive vs. Sophisticated Reasoning}

\begin{figure}[!h]
\centering
\includegraphics[width=0.99\columnwidth]{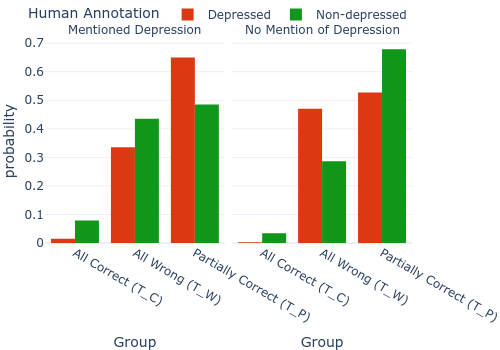}
\captionsetup{font=footnotesize}
\caption{Distribution of samples over the $T_C$, $T_P$, and $T_W$ collections w.r.t. different annotation groups.}
\label{fig:distri-tc-tp-tw}
\end{figure}

Given that revision of reasoning is possible in few-shot generation, we devise two schemes to select the qualified responses for LLM finetuning:
\begin{enumerate*}[label=(\arabic*)]
\item \textit{Intuitive reasoning (IR)} provides the desired responses with all correct predictions on the first attempt, without sophisticated logical analysis.
\item \textit{Sophisticated reasoning (SR)} permits a second attempt and allows for the revision of predictions.
\end{enumerate*}
In practice, the \textit{correct} \textit{IR} criteria filter the subtask predictions with their ``first'' labeling, while the \textit{SR} criteria use the ``last'' labeling, given that the majority of the responses provide only one prediction for a single subtask. 
Regarding all \textit{correct} responses $\{ R^{C}_{i,j} | i \in DepTweet \}$, there are 4,653 qualified responses collected as \textit{IR} and 4,649 that satisfy the criteria for \textit{SR}. Among them, six IR responses are replaced by two SR descriptions (one by Llama and the other by Qwen). This implies that only two generated diagnoses achieved \textit{correct} revisions.
Interestingly, the derived $T_C$, $T_P$, and $T_W$ sample collections are identical regardless of the IR and SR criteria. This suggests that these two samples are analyzed correctly by other LLMs on the first attempt. We visualize these three collections w.r.t. human annotation, as shown in Figure \ref{fig:distri-tc-tp-tw}. It is observed that the distribution of depressed samples with the presence of depression keywords is similar to the non-depressed without the keywords, and vice versa. This further indicates that these keywords influence the separation of the three groups.
Finally, we obtain 87 samples in the $T_C$ collection, 1963 in $T_P$, and 1082 in $T_W$.

\subsubsection{Performance Comparison}

We conduct instruction tuning on Llama and compare the qualification of reasoning under IR and SR criteria. Note that only a small number of different reasoning samples were included in the training data compared to the shared responses. The evaluation (on $T_W$) using few-shot learning as a baseline is shown in Table \ref{table:finetune} and \ref{table:finetune-bias}. Several observations can be made.
\begin{enumerate*}[label=(\arabic*)]
\item DPO outperforms SFT for the detection of subtasks and the joint correctness.
\item DPO$_{SR}$ outperforms DPO$_{IR}$ w.r.t. the joint \textit{correct ratio}, despite there being only six different responses. This might suggest that sophisticated reasoning has the potential to enhance DPO optimization.
\item The noticeable improvement in the \textit{correct ratio} of joint subtask predictions confirms the quality assurance of the responses used in the instruction tuning.
\item Both DPO and SFT reduce FP and FN rates for the \textit{MD} group. FP is also reduced at the cost of a higher FN rate for the \textit{NMD} group.
\end{enumerate*}

\begin{table}[!htbp]\tiny
\centering
\begin{tabular}{l|l|l|l|l|l}
\hline
LLM               & Baseline      & SFT$_{IR}$    & DPO$_{IR}$    & SFT$_{SR}$    & DPO$_{SR}$ \\ \hline
\textbf{A}        & \textbf{95.3} & 94.4          & 91.0          & 90.8          & 94.0 \\
\textbf{D} (F1)   & 75.0          & 79.4          & \textbf{83.5} & 75.8          & 81.7 \\
\textbf{S} (F1)   & 84.0          & 86.8          & 89.8          & 87.7          & \textbf{90.1} \\
\textbf{S1} (F1)  & 81.2          & 88.1          & 97.1          & 84.9          & \textbf{97.3} \\
\textbf{S2} (F1)  & 68.0          & 69.0          & 71.6          & 68.5          & \textbf{72.9} \\
\textbf{S3} (F1)  & 90.7          & 91.8          & \textbf{95.5} & 92.0          & 95.3 \\
\textbf{S4} (F1)  & 70.2          & 73.9          & 72.9          & \textbf{74.1} & 72.7	 \\
\textbf{S5} (F1)  & 96.2          & 96.3          & \textbf{97.5} & 96.5          & 97.5 \\
\textbf{S6} (F1)  & 85.9          & 87.8          & \textbf{89.8} & 86.8          & 89.2 \\
\textbf{S7} (F1)  & 93.2          & 96.3          & 98.6          & 96.4          & \textbf{99.0} \\
\textbf{S8} (F1)  & 91.9          & 93.0          & 93.6          & 92.7          & \textbf{93.7} \\
\textbf{S9} (F1)  & 93.8          & 93.8          & \textbf{94.6} & 93.7          & 94.3 \\ \hline
PHQ9 ($C$)        & 12.4          & 23.5	         & 32.3          & 23.0          & \textbf{33.4} \\ 
PHQ9+D ($C$)      & 8.0           & 20.1          & 27.9          & 19.7          & \textbf{29.8} \\
S+PHQ9+D ($C$)    & 0.0           & 12.0          & 22.0          & 12.8          & \textbf{23.7} \\ \hline
\end{tabular}
\captionsetup{font=footnotesize}
\caption{Evaluation of instruction tuning with Llama in comparison to the few-shot baseline. (\textbf{A} denotes \textit{adherence to the format}, \textbf{D} depression, and \textbf{S} \textit{speaker reference})}
\label{table:finetune}
\end{table}

\begin{table}[!htbp]\small
\centering
\begin{tabular}{l|cc|cc}
\hline
         & \multicolumn{2}{c|}{\textit{MD}} & \multicolumn{2}{c}{\textit{NMD}} \\ \hline
         & FP\%       & FN\%      & FP\%       & FN\%       \\ \hline
Baseline & 56         & 10        & 32         & 33         \\
SFT$\_IR$  & 47         & 5         & 13         & 39         \\
DPO$\_IR$  & 42         & 8         & 6          & 51         \\
SFT$\_SR$  & 54         & 5         & 22         & 34         \\
DPO$\_SR$  & 41         & 8         & 6          & 52 \\ \hline        
\end{tabular}
\captionsetup{font=footnotesize}
\caption{Comparison of detection weaknesses w.r.t. false position (FP) and false negative (FN) rates with instruction tuning. \textit{MD} refers \textit{Mentioned Depression}; \textit{NMD} denotes \textit{No Mention of Depression}. FP and FN are computed using the same methods as in Figure \ref{fig:sota-eval-md-nmd} and \ref{fig:fewshot-eval-md-nmd}. Note that DPO$\_SR$ is not distinct from DPO$\_IR$ w.r.t. single-task depression detection due to the trivial difference (two samples) in their traning datasets.}
\label{table:finetune-bias}
\end{table}

\section{Conclusion}

We systematically evaluated Mental-LLMs' performance with PHQ-9 labeling and investigated schemes for further optimization, arguing that single-task reasoning with proprietary LLMs may be insufficient for generalized depression detection.  
We proposed detailed subtask annotations and response qualification with reasoning. We illustrated that these subtask annotations implemented strict logical flows that not only represented ``good'' detection reasoning but also facilitated high-standard qualification for the selection of the tuning data. 
Although the LLMs performed well on individual subtasks, the joint decision-making with complete step-by-step analysis was still significantly more challenging and would require future investigation.
Additionally, a detailed analysis revealed that the detections were skewed towards depression when explicit depression-related keywords were present in the text while they were skewed towards non-depression when such keywords were absent. This implies that LLMs find it harder to analyze implicit language for depression.
To mitigate these two problems without the high cost of collecting descriptive annotations from human experts, we utilized machine-generated responses, separated into intuitive and sophisticated reasoning collections, and optimized Llama with each of them. 
The results indicated that DPO significantly enhanced the joint labeling of subtasks compared to SFT, while sophisticated reasoning might have more potential to improve the reasoning capabilities of LLMs. Further studies on larger-scale data collection would help to confirm our conclusions.

\bibliography{custom}

\end{document}